\documentclass[11pt]{article}

\usepackage[final]{ACL2023}

\usepackage{times}
\usepackage{latexsym}
\usepackage{multirow}
\usepackage[T1]{fontenc}
\usepackage{booktabs}
\usepackage[utf8]{inputenc}

\usepackage{microtype}
\usepackage{graphicx}
\usepackage{inconsolata}

\title{PersLitEval: Fine-grained Benchmark and Evaluation of LLMs on Persian Literature Questions}

\author{
Ruhallah Niazi$^{*}$ \qquad Faeze Ghorbanpour$^{*}$ \qquad Alexander Fraser \vspace{.2cm}\\
School of Computation, Information and Technology, TU Munich \\
Munich Center for Machine Learning (MCML)\\
\vspace{.1cm}
{\tt \small \{ruhallah.niazi, faeze.ghorbanpour\}@tum.de}\\
\vspace{.1cm}
{\small $^{*}$Equal contribution.}
}

\begin{document}
\maketitle
\begin{abstract}
Despite impressive multilingual capabilities, large language models (LLMs) remain poorly evaluated on literary knowledge in non-English languages. We introduce PersLitEval, a benchmark of 4,514 Persian literature multiple-choice questions across eight fine-grained categories spanning spelling, literary devices, grammar, vocabulary, word formation, and conceptual understanding, sourced from materials for the Konkur university entrance examination. We evaluate six LLMs across ten prompting strategies, revealing striking category-level disparities
across three tiers of task difficulty: models reach higher accuracy on conceptual similarity tasks but struggle with formal linguistic analysis, with spelling and word formation proving the hardest across all models. Prompting strategy has a significant impact on performance, with explained few-shot examples yielding the best results, particularly on formal linguistic categories. An error analysis identifies three failure modes: semantic comprehension gaps, formal linguistic knowledge gaps, and counting/enumeration errors, suggesting that different categories require different improvement strategies. ~\footnote{We release PersLitEval to support future research in Persian literary evaluation. Code is available at: 
\href{https://anonymous.4open.science/r/Fine-grained-benchmark-and-evaluation-of-LLMs-on-Persian-literature-Questions-340C/README.md}{anonymous.4open.science}.} 

\end{abstract}

\section{Introduction}

Large language models (LLMs) have achieved strong performance on a broad range of natural language understanding benchmarks, with evaluation efforts remaining heavily skewed toward English and general-purpose tasks \cite{hendrycks2020measuring, go2023compositional}. Multilingual benchmarks have expanded coverage to many languages, but typically assess surface-level capabilities, leaving domain-specific competencies, particularly literary knowledge, largely unevaluated \cite{ song2023globalbench,romanou2024include,xuan2025mmlu}. 

Persian is spoken by over 110 million people and has a literary tradition spanning more than a millennium, yet LLM competence in Persian literature remains largely unexamined. As LLMs are increasingly applied to literary analysis, computational humanities, and domain-specific NLP tasks \cite{kalhor2026ghazalbench, percul2025}, understanding their capabilities and failure modes across the full spectrum of literary knowledge becomes important, both for building reliable tools and for guiding future multilingual model development \citep{zhong2024opportunities}.

Existing Persian benchmarks have made important strides in general language understanding \cite{khashabi-etal-2021-parsinlu}, factual knowledge \cite{sakhaeirad2026unmasking}, and cultural alignment \cite{farsi2025melacmassiveevaluationlarge, percul2025}; But, we still lack a benchmark for \textit{Persian literary competence}. Literary knowledge in Persian goes beyond general reading comprehension or cultural familiarity. It requires mastery of several distinct linguistic skills, including rhetorical devices, grammar, spelling conventions, and word formation, as well as the ability to interpret meaning in texts. These skills are formally defined and represent different aspects of language competence, requiring evaluation of how LLMs handle this linguistic knowledge and where they succeed or fail.


We introduce \textbf{PersLitEval}, a benchmark of 4,514 multiple-choice questions across eight categories of Persian literary knowledge, sourced from domain-expert-authored Konkur preparation materials. We evaluate six LLMs under ten prompting strategies, and conduct a category-level error analysis on Gemini 2.5 Flash \citep{comanici2025gemini}, the best-performing model. Our main contributions are: \textbf{(1)} the first benchmark specifically targeting Persian literary knowledge as formally assessed in Persian education, covering both formal linguistic analysis and conceptual understanding; \textbf{(2)} a systematic evaluation of six LLMs spanning both open-source and closed-source families across ten prompting strategies, revealing that models perform better on conceptual tasks such as semantic relatedness and main message than on surface form tasks such as spelling and word formation and that explained few-shot prompting yields the best performance which demonstrate explained examples help activate knowledge, \textbf{(3)} a fine-grained error analysis that identifies three recurring failure modes: semantic comprehension gaps, formal linguistic knowledge gaps, and counting or enumeration errors, offering a clearer picture of where current LLMs struggle on structured literary tasks.

\section{Related Work}
Compared to existent benchmarks on Persian language, such as ParsiNLU \cite{khashabi-etal-2021-parsinlu} covering broad natural language understanding tasks, Khayyam Challenge \cite{ghahroodi2024khayyamchallengepersianmmlullm} adapting MMLU-style language understanding and evaluation, FarsEval \cite{shamsfard2025farseval} spanning medicine, law, and cultural knowledge, MELAC \cite{farsi2025melacmassiveevaluationlarge} evaluating cultural alignment, and \citet{abaskohi2024benchmarkinglargelanguagemodels} benchmarking general Persian NLP tasks, ours is a benchmark focused exclusively on \textit{Persian literary knowledge}, covering a taxonomy of formally codified skills that define literary knowledge in formal Persian education.
Most closely related, \citet{tourajmehr2025evaluating} evaluates LLMs on generating creative Persian literary texts across four creativity dimensions, and \citet{nobakhtian2025evaluating} evaluates recognizing classical poetry allusions; while both address Persian literary phenomena, neither evaluates the broad formal, conceptual, or linguistic skills.



\section{PersLitEval Benchmark} 
\textbf{Data Source and Curation}\
PersLitEval consists of 4,514 four-choice Persian literature questions sourced from publicly available Konkur preparation documents downloaded from Konkur.in \footnote{\url{http://konkur.in/}}, a widely used repository of Iranian university entrance examination resources. Questions were extracted from documents using Google Cloud OCR\footnote{\url{https://cloud.google.com/}}, followed by normalization
and cleaning. A subset of 490 questions includes expert-written answer explanations, used in our explained few-shot prompting experiments. More details are provided in \ref{sec:dataset}. 

\noindent
\textbf{Categories}\
Each question was assigned to one of eight categories based on its content and required skill. \textit{Semantic relatedness} tests whether two texts share conceptual meaning; \textit{main message} tests identification of a passage's central idea; \textit{vocabulary} targets precise word meaning in classical context; \textit{literary devices} requires recognition of classical rhetorical devices; \textit{grammar} covers syntactic analysis and sentence structure; \textit{word formation} examines morphological patterns; \textit{spelling} tests correct orthographic forms of vocabulary. An \textit{Others} category includes questions across multiple skills that do not fit into the other categories. The distribution of categories is provided in Table \ref{tab:categories}.

\begin{table}
\centering
\begin{tabular}{lcc}
\hline
\textbf{Category} & \textbf{All} & \textbf{Expl.} \\
\hline
Main message & 1,107 & 112 \\
Semantic relatedness & 1,031 & 80 \\
Grammar & 927 & 55 \\
Others & 676 & 62 \\
Word formation & 430 & 15 \\
Literary devices & 151 & 84 \\
Vocabulary & 104 & 26 \\
Spelling & 88 & 56 \\
\hline
\textbf{Total} & \textbf{4,514} & \textbf{490} \\
\hline
\end{tabular}
\caption{Distribution of questions across categories (full dataset vs.\ subset with explanations).}
\label{tab:categories}
\end{table}



\section{Experimental Setup}

\begin{table*}[t]
\centering
\small
\setlength{\tabcolsep}{4pt}
\renewcommand{\arraystretch}{0.88}
\begin{tabular}{llcccccccccc}
\hline
\textbf{Tier} & \textbf{Category} & \textbf{ZS} & \textbf{SJ} & \textbf{CoT} & \textbf{TR} & \textbf{DEF} & \textbf{CAU} & \textbf{FS} & \textbf{FS+J} & \textbf{EFS} & \textbf{EFS+J} \\
\hline
\multirow{3}{*}{\textit{T1: Conceptual}}
  & Sem.\ Relatedness & 67.41 & 69.35 & 65.47 & 65.47 & —     & 67.99 & \textbf{72.76} & 71.98 & 71.19 & 71.87 \\
  & Main Message      & 58.45 & 59.98 & 56.64 & 56.55 & —     & 62.06 & 66.03 & 66.49 & \textbf{66.58} & 64.05 \\
  & Vocabulary        & 57.69 & 57.69 & 53.85 & 50.96 & —     & 49.04 & 54.46 & 51.49 & 57.69 & \textbf{58.65} \\
\hline
\multirow{2}{*}{\textit{T2: Formal Ling.}}
  & Literary Devices  & 28.48 & 33.77 & 33.77 & 33.11 & 40.40 & 40.40 & 39.86 & 41.89 & 43.05 & \textbf{46.36} \\
  & Grammar           & 40.45 & 42.83 & 41.53 & 39.27 & 43.26 & 41.75 & 43.83 & 43.40 & \textbf{45.63} & 43.80 \\
\hline
\multirow{2}{*}{\textit{T3: Surface Form}}
  & Spelling          & 25.00 & 29.55 & 36.36 & 27.27 & —     & 29.55 & 31.76 & 29.41 & \textbf{40.91} & 37.50 \\
  & Word Formation    & 36.98 & 39.30 & 35.81 & 37.67 & 35.58 & 37.21 & \textbf{39.34} & 37.94 & 37.91 & 36.51 \\
\hline
\textit{Other}
  & Others           & 46.13 & 45.95 & 43.42 & 45.05 & —     & 45.23 & 44.57 & 46.38 & \textbf{48.47} & 47.03 \\
\hline
\end{tabular}
\caption{\small{Gemini 2.5 Flash accuracy (\%) by prompting strategy and category (bold = best per row).
ZS=Zero-shot, SJ=Short justification, CoT=Chain-of-thought, TR=Translation,
DEF=Definitions (applied only where literary term definitions exist; —\,=\,N/A),
CAU=Caution, FS=Few-shot,
EFS=Explained few-shot, +J=+justification.}}
\label{tab:gemini_prompts}
\end{table*}

\textbf{Models} \
We evaluate six LLMs spanning five major labs: Gemini 2.5 Flash \citep{comanici2025gemini}, GPT-4o and GPT-4.1 Mini \citep{openai2024gpt4technicalreport, openai2024gpt4ocard, openai_gpt4_2023}, Grok 4 Fast \citep{xai_grok4_fast_2025}, Llama 3.3 70B Instruct \citep{touvron2024llama3}, and Qwen2.5 72B Instruct \cite{qwen2_2024}, covering both closed-source commercial and open-source models. For more details, see Appendix \ref{sec:models}.

\noindent
\textbf{Prompting Strategies} \
We evaluate ten prompting strategies to examine how prompt design affects performance on Persian literary tasks. \textit{Zero-shot} provides the question directly with no additional guidance. \textit{Short justification} asks the model to briefly justify its answer. \textit{Chain-of-thought} (CoT) instructs the model to reason step by step before answering. \textit{Few-shot} provides three randomly sampled in-category examples without explanations. \textit{Translation} instructs the model to internally translate the question into English before answering. \textit{Definitions} provides definitions of relevant literary terms alongside the question. \textit{Caution} supplies per-category guidance derived from error analysis, warning the model against known failure modes such as overcounting or inventing spelling errors. \textit{Explained few-shot} provides three randomly sampled examples from the subset of questions containing expert-written answer explanations. We have \textit{+ justification} prompts that extend each few-shot strategy by asking the model to justify its answer. All instruction prompts were written in English, while questions and answer choices were presented in Persian. Full prompt texts are in Appendix \ref{sec:prompts}.


\section{Evaluation}


Table~\ref{tab:gemini_prompts} presents Gemini 2.5 Flash accuracy across all prompting strategies and categories grouped by tier, revealing two consistent patterns. First, a clear model hierarchy emerges. Gemini 2.5 Flash (45.07\%) and GPT-4o (41.98\%) substantially outperform the remaining four models. The stronger performance of the top two models likely reflects greater exposure to Persian literary content during pretraining. GPT-4.1 Mini, Llama 3.3 70B, and Qwen2.5 72B cluster between 32--35\%, while Grok 4 Fast scores lowest at 30.39\%. Despite differences in scale, training objectives, and language coverage, their performance converges to a similar range, suggesting that none of these factors alone is sufficient for Persian literary knowledge.

\begin{figure*}
\centering

\includegraphics[width=0.85\linewidth,page=1]{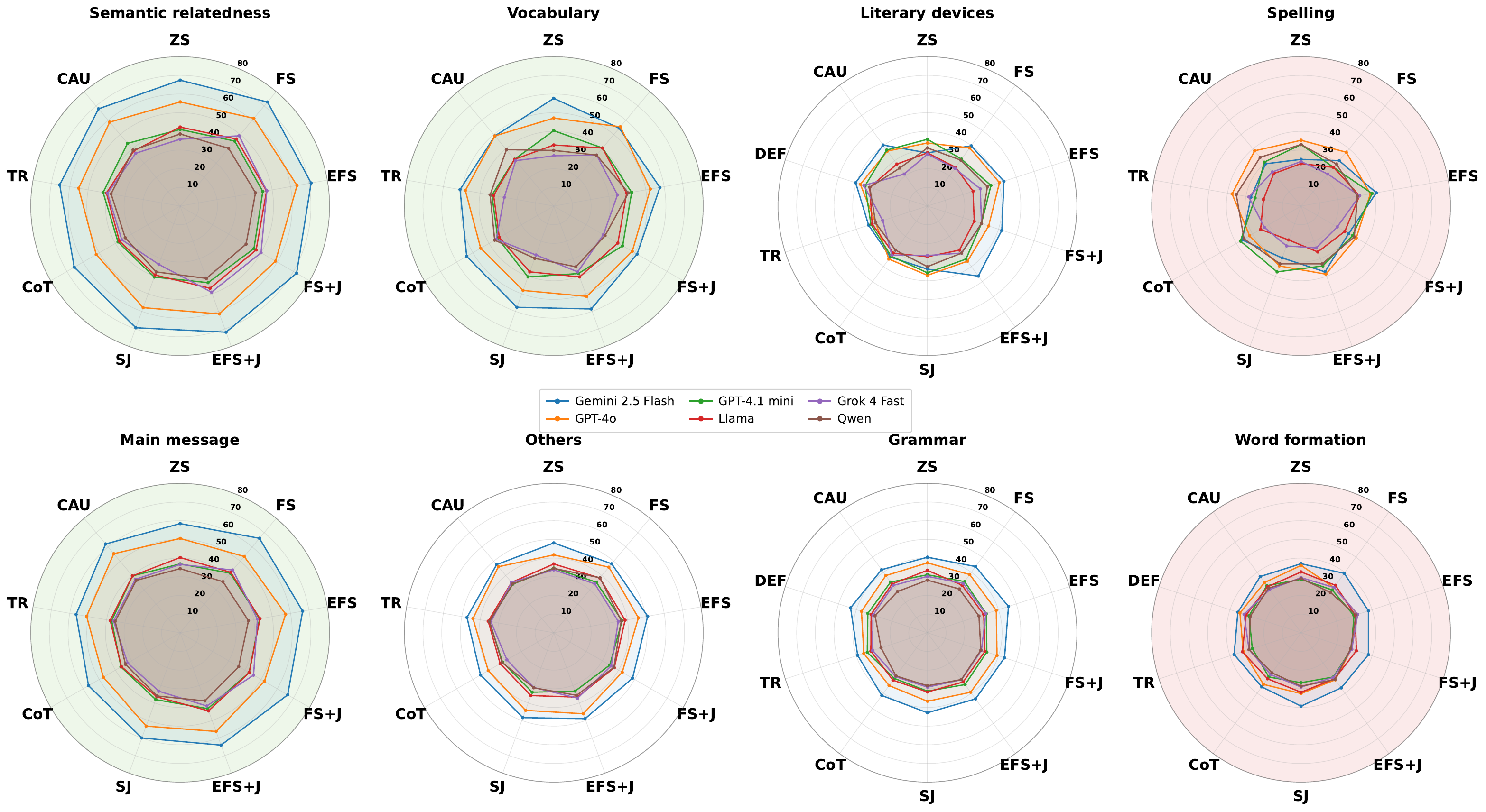}

\caption{Radar plots showing accuracy of all models across all prompting strategies per category. }
\label{fig:radar}
\end{figure*}

Second, category difficulty follows a clear three-tier pattern. Conceptual categories consistently yield the highest scores across all models, with semantic relatedness reaching 67.41\% for Gemini zero-shot and 72.76\% under the best prompt. Formal linguistic categories fall in the mid range (32.18\% cross-model average), while surface form categories prove hardest — Gemini's spelling score of 25.00\% zero-shot sits exactly at random baseline, rising to only 40.91\% under the best prompt. This gap is not model-specific: it holds across all six models and all prompting strategies, pointing to a systematic difference in how Persian literary knowledge is distributed across task types in current LLM training data.
Regarding prompting, explained few-shot and standard few-shot achieve the highest overall averages (42\% each), while translation and CoT rank lowest (38\% each). The effect of prompting varies substantially by category tier. CoT, for instance, shows its largest gain on spelling while hurting conceptual categories. Figure~\ref{fig:radar} shows radar plots for all models across prompting strategies by category. 
We discuss results, underlying reasons, and error patterns by tier. Error analysis and positional bias experiment are in Appendices~\ref{sec:error} and~\ref{sec:bias}.





\noindent
\textbf{Tier 1; Conceptual Categories.}
Semantic relatedness, main message, and vocabulary are the strongest categories across all models. Gemini leads with GPT-4o following it, while the remaining models cluster with lower accuracy.
The relatively strong performance in these categories can be attributed to two factors: LLMs are widely pretrained on Persian prose, including news, web text, and social media, providing sufficient exposure for meaning-level reasoning, and conceptual similarity tasks structurally resemble reading comprehension, a well-studied capability that partially transfers to Persian. Few-shot and explained few-shot yield the best results, while CoT consistently underperforms zero-shot across all models. Translation prompting performs worst across all models on these categories, confirming that Persian poetic meaning is culturally embedded and cannot be recovered by routing through English. 

Error analysis reveals that 84--90\% of mistakes in semantic relatedness and main message are wrong meaning errors; models grasp broad poetic context but fail on fine-grained distinctions such as differentiating preference from negation or cause from result in classical verse. In vocabulary, 68\% of errors are wrong meaning and an additional 16\% involve wrong semantic relation, suggesting surface-level lexical association rather than deep classical vocabulary knowledge.

\noindent
\textbf{Tier 2; Formal Linguistic Categories.}
Literary devices and grammar prove substantially harder.
These categories suffer from a compounded knowledge gap: formal Persian literary rules and grammar terminology are highly specialized and severely underrepresented in general pretraining corpora, literary devices require textbook-precise definitions rather than approximate understanding, and visually similar devices, such as simile and metaphor, or personification and allegory, are easily confused without rigorous formal training. A notable observation is that on literary devices, Gemini's zero-shot falls below GPT-4.1 Mini, yet recovers strongly with explained few-shot, suggesting that demonstrated examples effectively inject the missing formal knowledge. Grammar shows more uniform behavior across models, with definitions and explained few-shot providing consistent gains. 

Error analysis confirms that 79\% of literary device errors involve wrong device identification; models misapply textbook definitions or confuse similar devices; and 82\% of grammar errors involve wrong syntactic analysis, particularly on multi-step parsing tasks requiring several grammatical decisions simultaneously.

\noindent
\textbf{Tier 3; Surface Form Categories.}
Spelling and word formation are the hardest categories,
with cross-model averages of 28.8\% and 32.1\%; closest to the 25\% random baseline. Three factors explain this: first, most LLM tokenizers are trained predominantly on Latin-script data and produce suboptimal segmentation of Persian script, making character-level precision unreliable; second, word formation requires identifying morpheme boundaries, prefixes, suffixes, and compound structures, that subword tokenization does not respect in Persian; third, counting and enumeration tasks expose a known weakness of autoregressive LLMs, where models identify correct elements but systematically miscount them. Spelling is the one category where Gemini performs worse than GPT-4o and GPT-4.1 Mini, suggesting model-specific differences in Persian orthographic knowledge. CoT shows its largest gains in spelling, consistent with deliberate reasoning counteracting hasty hallucination, while word formation shows the narrowest prompt sensitivity, pointing to a hard ceiling that prompt engineering alone cannot overcome. 

Error analysis identifies two failure modes operating across both categories: in spelling, 48\% of errors are hallucinated mistakes that do not exist in the original text, and 26\% are counting errors; in word formation, 44\% involve wrong morphological analysis, and 38\% are counting failures, making Tier~3 the most resistant to prompt improvement.

\section{Conclusion}

We introduced \textbf{PersLitEval}, a fine-grained benchmark targeting Persian literary knowledge, and systematically evaluated six LLMs across ten prompting strategies. Our results reveal a clear three-tier difficulty pattern across categories, identify explained few-shot as the most effective prompting strategy, and expose three recurring failure modes through error analysis, offering a diagnostic foundation for future Persian LLM development. 


\section*{Limitations}

This work has several limitations. First, our evaluation is limited to multiple-choice questions, which may not capture the full range of Persian literary knowledge; open-ended generation and explanation tasks remain unexplored. Second, we do not include a human performance baseline, making it difficult to contextualize model scores against expert-level competence. Third, due to computational constraints, we were unable to evaluate additional frontier models, which may perform differently from those tested. Fourth, the benchmark is sourced exclusively from Konkur preparation materials, which may not represent the full diversity of Persian literary knowledge beyond the exam context. Finally, our error analysis is based on a sample of incorrect predictions from a single model and prompt condition, and may not generalize across all models and settings.

\section*{Ethical Considerations}
The dataset is sourced from publicly available Konkur preparation materials hosted on konkur.in, and does not contain any personally identifiable information. We contacted the website owner directly and received explicit permission via email to use the materials for research purposes; we have also committed to sharing our results with them. All models were accessed via API and evaluated solely on multiple-choice questions with no potential for harmful content generation. The benchmark is intended to support research in Persian NLP and multilingual LLM evaluation, and we foresee no direct negative societal impacts from its release.

\bibliography{custom}
\bibliographystyle{acl_natbib}

\appendix

\section{Dataset}
\label{sec:dataset}

 Across our source documents, we extracted 4,601 questions in total. 
 We then removed 87 questions that depended on visual structure, such as arrows, diagrams, tables, or spatial layout, because these could not be represented reliably in plain text. This left a final benchmark of 4,514 questions.

The extraction pipeline combined OCR-based processing with manual cleaning. We used the Google Cloud Vision API with the \texttt{DOCUMENT\_TEXT\_DETECTION} setting to extract dense text from the PDFs. Because the OCR output was optimized for Arabic-script text more generally, we normalized visually similar Arabic characters to their Persian equivalents in a post-processing step. Each retained question was then stored as a JSON object containing the question text, four answer choices, and the correct answer. In addition, 490 questions contain answer explanations, which were later used for the explained few-shot prompting conditions.

For categorization, we first applied a rule-based approach using category-specific keywords and question patterns, followed by manual review to correct potential errors. While the final categorization is not perfect, our error analysis suggests miscategorization is minimal, where approximately 5\% of questions were found to belong to other categories.
The \textit{Others} category contains heterogeneous questions that do not fit cleanly into one of the seven main categories. In many cases, these questions combine multiple skills at once, or their main requirement is too broad to justify a separate fine-grained label. We grouped them under \textit{Others} instead of creating many very small categories, since such categories would have been too sparse for meaningful comparison across models and prompting strategies.
Table~\ref{tab:dataset_examples} provides one example question for each category.

\begin{table}[t]
\centering
\includegraphics[width=\columnwidth]{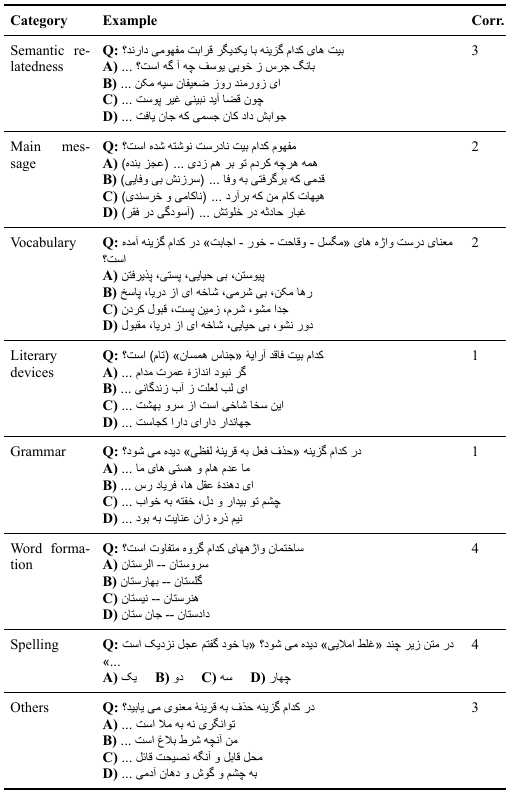}
\caption{One example question from each category.}
\label{tab:dataset_examples}
\end{table}

\paragraph{Contamination Risk.} The source materials are publicly available on konkur.in, so we cannot fully exclude the possibility that some questions appear in model pretraining data. However, the risk is mitigated by two factors: the materials are in Persian, a language underrepresented in most LLM pretraining corpora, and the original format is PDF with mixed text and images, requiring non-trivial extraction to obtain clean question text. We therefore consider contamination risk low but not negligible, and note that the consistently low performance across all models is inconsistent with systematic memorization.

\newpage
\section{Models}
\label{sec:models}
We evaluated six large language models accessed via API through OpenRouter, selected to represent diverse model families and providers. Table~\ref{tab:models} lists the model names and their OpenRouter identifiers. We set the temperature to 0, used deterministic hyperparameters, and fixed a random seed to ensure reproducibility. Due to budget and computational constraints, each experiment was run once, and we were unable to perform multiple runs with different random seeds.
\begin{table}[h]
\centering
\small
\begin{tabular}{lp{4cm}}
\hline
\textbf{Model} & \textbf{OpenRouter ID} \\
\hline
Gemini 2.5 Flash & \texttt{google/ gemini-2.5-flash} \\
GPT-4o & \texttt{openai/gpt-4o} \\
GPT-4.1 Mini & \texttt{openai/gpt-4.1-mini} \\
Grok 4 Fast & \texttt{x-ai/grok-4.1-fast:nitro} \\
Llama 3.3 70B Instruct & \texttt{meta-llama/ llama-3.3-70b-instruct} \\
Qwen2.5 72B Instruct & \texttt{qwen/ qwen-2.5-72b-instruct} \\
\hline
\end{tabular}
\caption{Evaluated models.}
\label{tab:models}
\end{table}

\section{Prompts}
\label{sec:prompts}

All instruction prompts were written in English, while questions and answer choices were presented in Persian. We evaluated ten prompting strategies in total, including zero-shot, short justification, chain-of-thought, translation, definitions, caution, few-shot, few-shot + justification, explained few-shot, and explained few-shot + justification. For standard few-shot strategies, three examples were randomly drawn from the same category as the test question. Explained few-shot used three randomly sampled examples from the subset of 490 questions with expert-written answer explanations.\\


\noindent\textbf{Zero-shot}
\begin{quote}\footnotesize\ttfamily
You will see a multiple-choice question in Persian about Persian literature.\\
\{\{TRANSLATION\_MODE\_INSTRUCTION\}\}\\[0.2em]
\{\{CATEGORY\_CAUTION\}\}\\
\{\{CATEGORY\_DEFINITIONS\}\}\\[0.2em]
FORMAT:\\
- Output ONLY one letter: A, B, C, or D.\\[0.2em]
RULES:\\
- Output exactly one line and nothing else.\\
- Do not give any explanation.\\
- Choose exactly one answer and stay consistent.\\
- Start directly with the chosen letter.
\end{quote}

\noindent\textbf{Zero-shot + justification}
\begin{quote}\footnotesize\ttfamily
You will see a multiple-choice question in Persian about Persian literature.\\
\{\{TRANSLATION\_MODE\_INSTRUCTION\}\}\\[0.2em]
\{\{CATEGORY\_CAUTION\}\}\\
\{\{CATEGORY\_DEFINITIONS\}\}\\[0.2em]
FORMAT:\\
- Line 1: output ONLY one letter: A, B, C, or D.\\
- Then you MAY write one very short justification in \{\{REASONING\_LANGUAGE\}\}.\\[0.2em]
RULES:\\
- Choose exactly one answer and stay consistent.\\
- Start directly with the chosen letter.
\end{quote}

\noindent\textbf{Zero-shot + chain-of-thought}
\begin{quote}\footnotesize\ttfamily
You will see a multiple-choice question in Persian about Persian literature.\\
\{\{TRANSLATION\_MODE\_INSTRUCTION\}\}\\[0.2em]
Think step by step in \{\{REASONING\_LANGUAGE\}\}.\\[0.2em]
\{\{CATEGORY\_CAUTION\}\}\\
\{\{CATEGORY\_DEFINITIONS\}\}\\[0.2em]
Steps:\\
- Restate the meaning briefly.\\
- Identify the main point.\\
- Eliminate mismatches.\\
- Choose the best answer.\\[0.2em]
FORMAT:\\
- Line 1: output ONLY one letter: A, B, C, or D.\\
- Then write step-by-step reasoning in \{\{REASONING\_LANGUAGE\}\}.\\[0.2em]
RULES:\\
- Choose exactly one answer and stay consistent.\\
- Start directly with the chosen letter.
\end{quote}

\noindent\textbf{Few-shot}
\begin{quote}\footnotesize\ttfamily
You will be given a multiple-choice question about Persian literature.\\
\{\{TRANSLATION\_MODE\_INSTRUCTION\}\}\\[0.2em]
FORMAT:\\
- Output ONLY one letter: A, B, C, or D.\\[0.2em]
\{\{CATEGORY\_CAUTION\}\}\\
\{\{CATEGORY\_DEFINITIONS\}\}\\[0.2em]
Examples:\\
\{\{FEWSHOTS\}\}\\
-- End of examples --\\[0.2em]
RULES:\\
- Do not give any explanation.\\
- Choose exactly one answer and stay consistent.\\
- Start directly with the chosen letter.
\end{quote}

\noindent\textbf{Few-shot + justification}
\begin{quote}\footnotesize\ttfamily
You will be given a multiple-choice question about Persian literature.\\
\{\{TRANSLATION\_MODE\_INSTRUCTION\}\}\\[0.2em]
FORMAT:\\
- Output ONLY one letter: A, B, C, or D.\\
- Then write a short explanation in \{\{REASONING\_LANGUAGE\}\}.\\[0.2em]
\{\{CATEGORY\_CAUTION\}\}\\
\{\{CATEGORY\_DEFINITIONS\}\}\\[0.2em]
Examples:\\
\{\{FEWSHOTS\}\}\\[0.2em]
RULES:\\
- Choose exactly one answer and stay consistent.\\
- Start directly with the chosen letter.
\end{quote}

\noindent
Translation, caution, and definitions were instantiated through the corresponding placeholders. Explained few-shot used the same few-shot templates, but replaced \texttt{\{\{FEWSHOTS\}\}} with examples that additionally contained expert-written explanations. The base templates and category-specific guidance below follow the prompt configuration used in the code.

\begin{center}
\begin{minipage}{\columnwidth}
\scriptsize
\setlength{\tabcolsep}{3pt}
\renewcommand{\arraystretch}{1.0}
\begin{tabular}{p{1.8cm}p{5.1cm}}
\hline
\textbf{Category} & \textbf{Inserted guidance} \\
\hline
Semantic relatedness & Avoid superficial keyword overlap; focus on exact semantic distinctions. \\
Main message & Focus only on the main claim; watch negation, exception, and close distractors. \\
Vocabulary & Use precise exam-oriented meaning; avoid vague guesses from form or root similarity. \\
Word formation & Use word-formation guidance; when triggered, insert definitions for simple, derived, compound, derived-compound, prefix, suffix, ezafe compound, and descriptive compound. \\
Grammar & Use textbook grammar guidance; when triggered, insert definitions for subject, predicate, copular verb, object, prepositional complement, adverbial, ezafe, and dependent/dependent-of-dependent. \\
Literary devices & Use textbook literary-device guidance; when triggered, insert definitions for simile, metaphor, majaz, kenayah, eiham, eiham tanasub, jenas, semantic field harmony, antithesis, and talmih. \\
Spelling & Do not invent spelling errors; count only certainly incorrect items. \\
Others & Read all options fully; do not rely on superficial overlap or invent missing information. \\
\hline
\end{tabular}
\captionsetup{type=table}
\captionof{table}{Category-specific guidance and definitions.}
\label{tab:category_guidance}
\end{minipage}
\end{center}

\section{Positional Bias}
\label{sec:bias}
Positional bias means that a model prefers certain answer options more often than others, independent of question content. This matters in multiple-choice evaluation because such preferences can distort accuracy. To inspect this, we analyzed how often each of the six evaluated models selected options A, B, C, and D across the evaluation set. 

The closed-source models did not show noticeable positional bias. As shown in Figure~\ref{fig:bias_closed}, Gemini 2.5 Flash, GPT-4o, GPT-4.1 Mini, and Grok 4 Fast produced relatively balanced answer-option distributions. The two open-source models showed mild residual preferences. As shown in Figure~\ref{fig:bias_open}, Qwen2.5 72B Instruct leaned slightly toward option B, while Llama 3.3 70B Instruct leaned slightly toward option C.

\begin{center}
\begin{minipage}{\columnwidth}
\captionsetup{type=figure}
\centering
\begin{minipage}[t]{0.49\columnwidth}
    \centering
    \includegraphics[width=\textwidth]{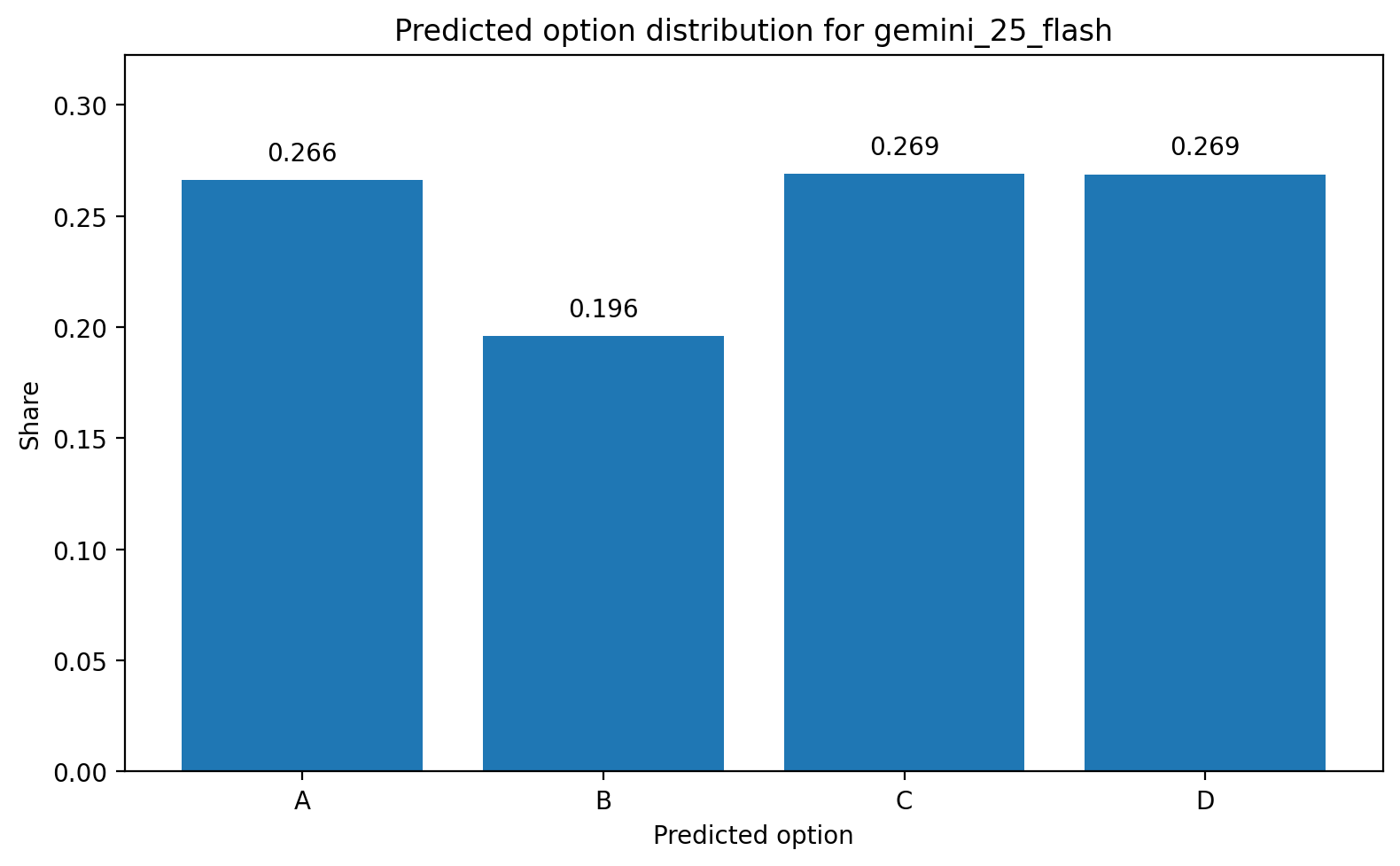}
\end{minipage}
\hfill
\begin{minipage}[t]{0.49\columnwidth}
    \centering
    \includegraphics[width=\textwidth]{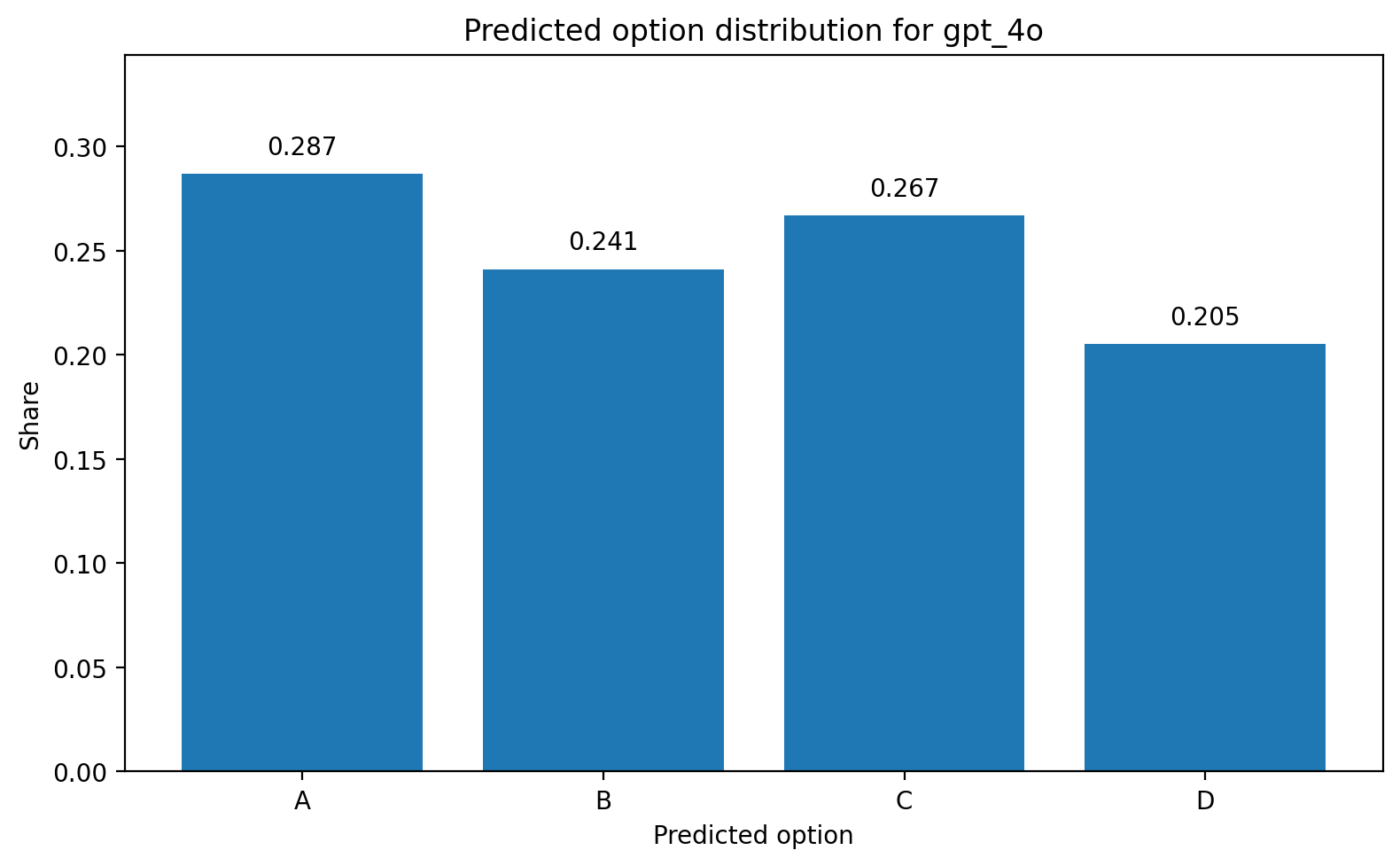}
\end{minipage}

\vspace{0.4em}

\begin{minipage}[t]{0.49\columnwidth}
    \centering
    \includegraphics[width=\textwidth]{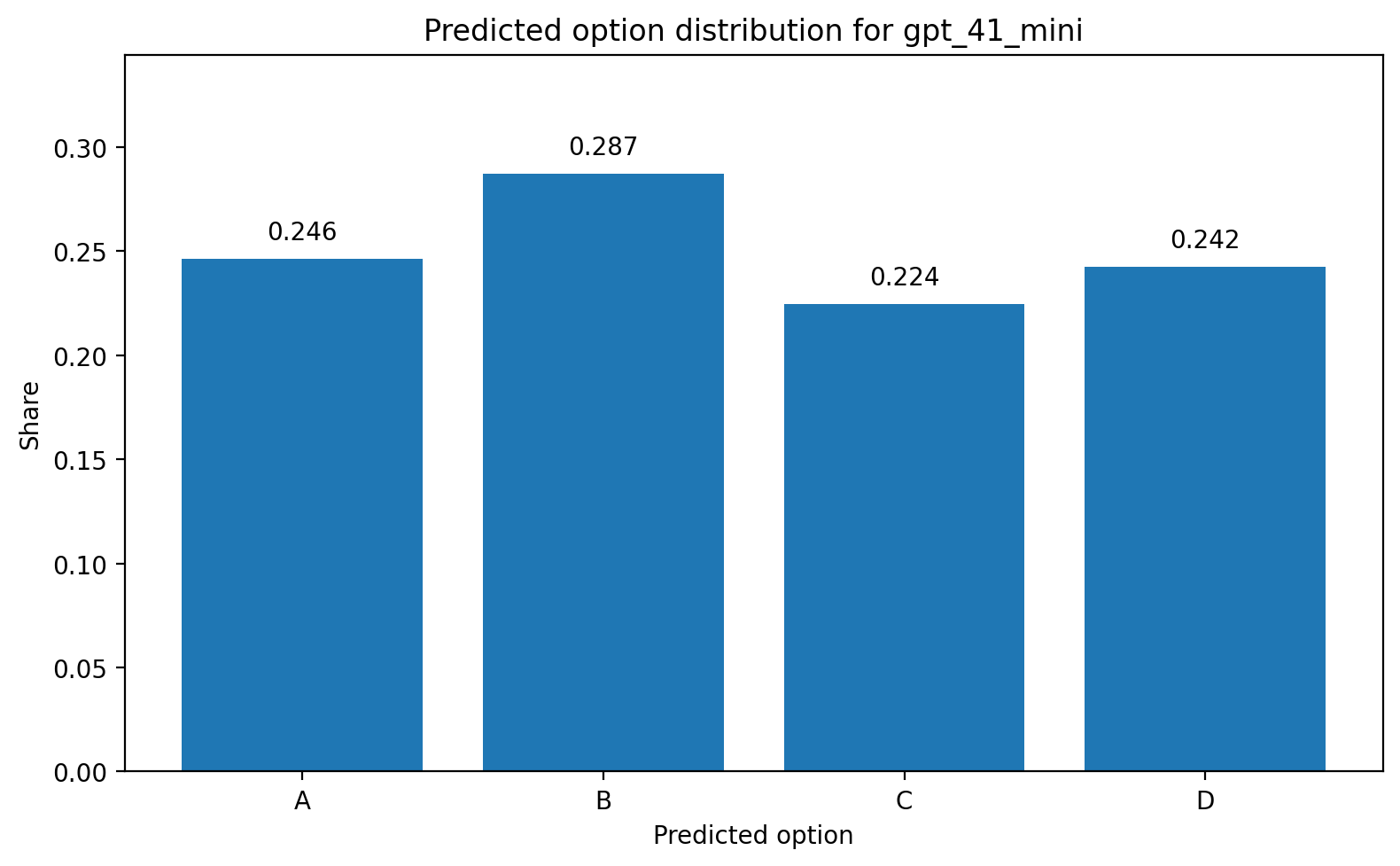}
\end{minipage}
\hfill
\begin{minipage}[t]{0.49\columnwidth}
    \centering
    \includegraphics[width=\textwidth]{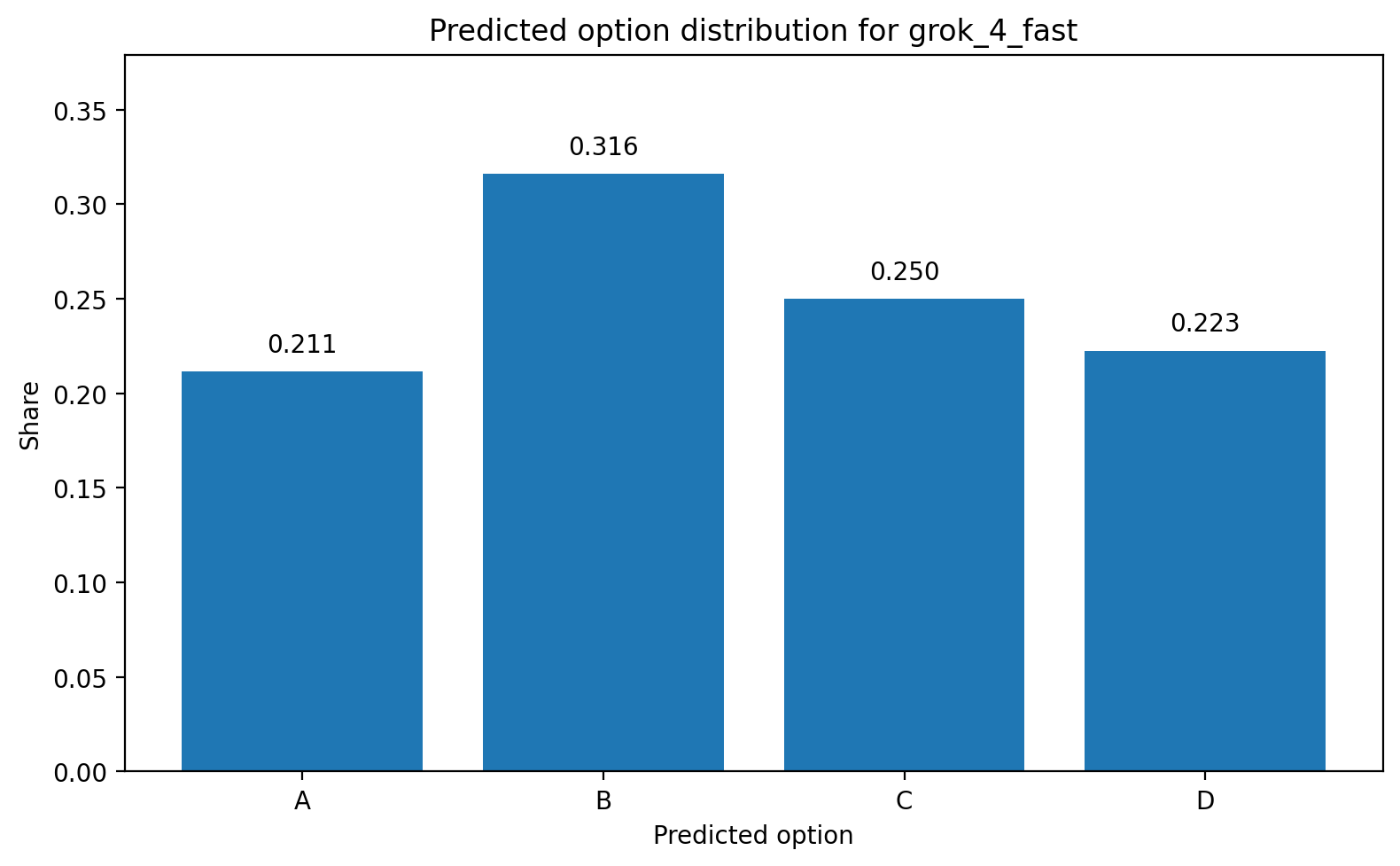}
\end{minipage}

\captionof{figure}{Predicted answer-option distributions for Gemini 2.5 Flash, GPT-4o, GPT-4.1 Mini, and Grok 4 Fast.}
\label{fig:bias_closed}
\end{minipage}
\end{center}

\begin{center}
\begin{minipage}{\columnwidth}
\captionsetup{type=figure}
\centering
\begin{minipage}[t]{0.49\columnwidth}
    \centering
    \includegraphics[width=\textwidth]{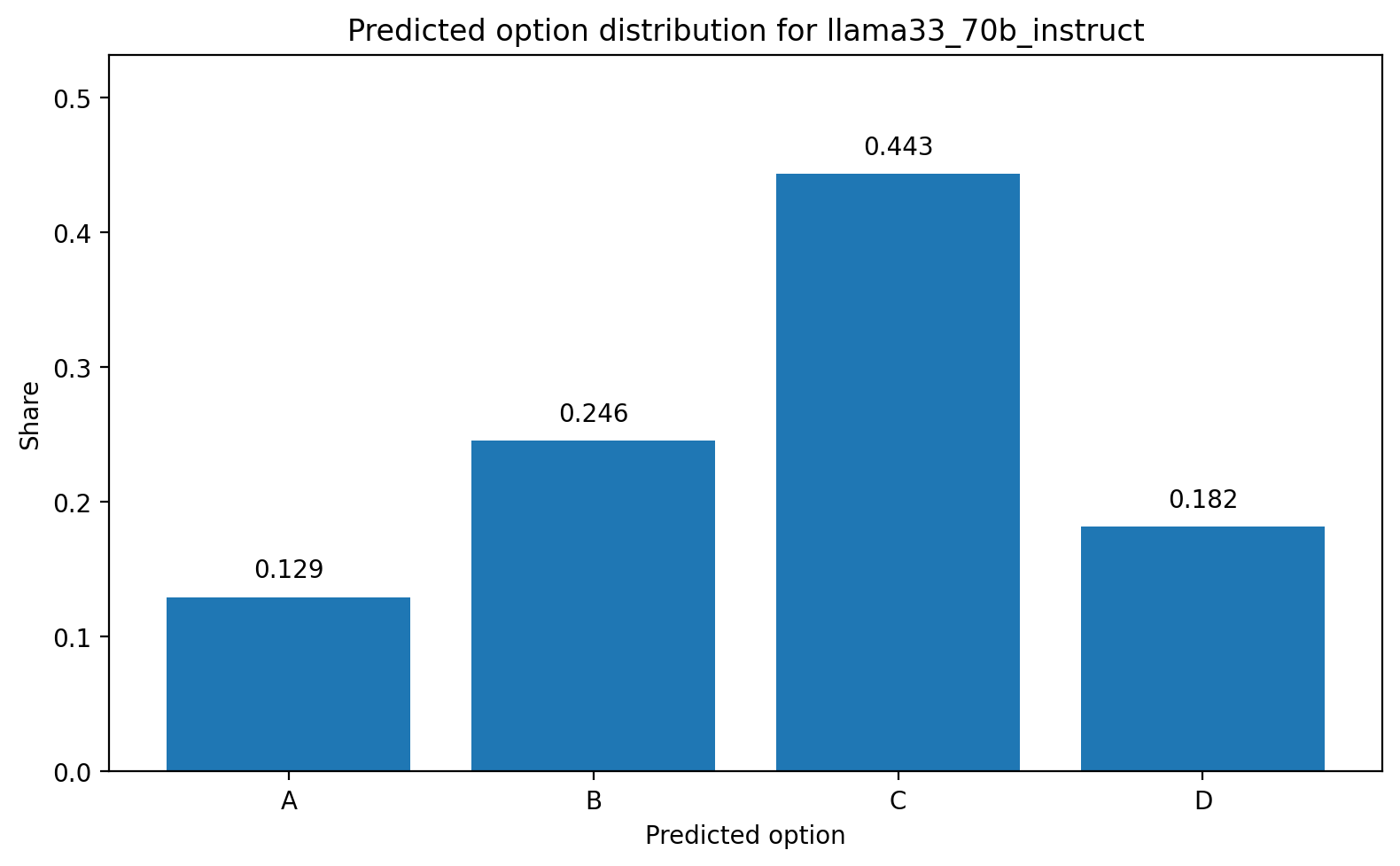}
\end{minipage}
\hfill
\begin{minipage}[t]{0.49\columnwidth}
    \centering
    \includegraphics[width=\textwidth]{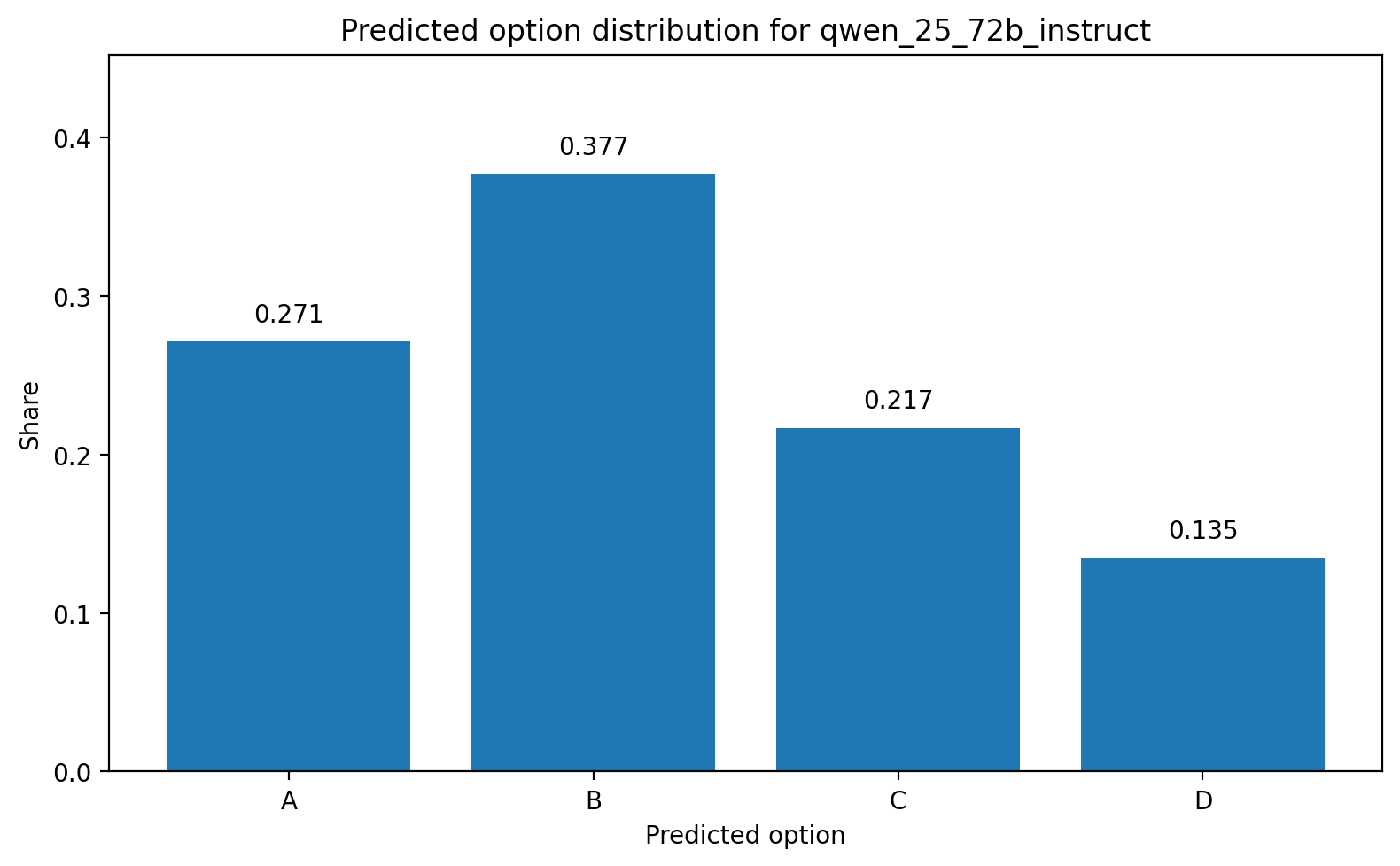}
\end{minipage}
\captionof{figure}{Predicted answer-option distributions for Llama 3.3 70B Instruct and Qwen2.5 72B Instruct.}
\label{fig:bias_open}
\end{minipage}
\end{center}

Overall, the six evaluated models were usable for multiple-choice evaluation, but the closed-source models appeared more balanced across answer positions than the two open-source models.


\section{Error Analysis Details}
\label{sec:error}
 We analyzed incorrect predictions from Gemini 2.5 Flash under the explained few-shot + justification prompt. The sampling strategy depended on the size of the category. For larger categories, we sampled 50 incorrectly answered questions. For smaller categories, we analyzed all available incorrect answers. For example, although the Literary devices category had more than 50 errors, the category itself was small enough to be inspected fully, so all 87 incorrect predictions were analyzed.
\paragraph{Error taxonomy.}
Each incorrect prediction was assigned to one dominant error type. \textit{Wrong meaning} refers to cases where the model misunderstood the intended meaning of a verse, passage, or answer option. \textit{Wrong relation} refers to cases where the model relied on superficial similarity instead of the intended semantic relation. \textit{Wrong literary device} refers to incorrect identification of rhetorical devices. \textit{Wrong grammar analysis} refers to errors in syntactic or grammatical interpretation. \textit{Wrong word-formation analysis} refers to incorrect analysis of morphemes, compounds, or derivational structure. \textit{Wrong spelling} refers to cases where the model failed to identify the correct spelling error or hallucinated an error. \textit{Wrong counting} refers to cases where the model identified relevant elements but counted them incorrectly. \textit{Wrong factual knowledge} refers to cases where the model lacked the required background knowledge. Finally, \textit{wrong category} refers to cases where the question belonged to a different category.

\begin{table}[t]
\centering
\small
\setlength{\tabcolsep}{3pt}
\renewcommand{\arraystretch}{1.05}
\begin{tabular}{p{2.0cm}p{4.8cm}}
\hline
\textbf{Category} & \textbf{Error distribution} \\
\hline
Semantic relatedness & Wrong meaning: 45; wrong category: 5 \\
Main message & Wrong meaning: 42; wrong category: 8 \\
Vocabulary & Wrong meaning: 30; wrong relation: 7; wrong counting: 5; wrong category: 2 \\
Literary devices & Wrong literary device: 69; wrong category: 16; wrong counting: 2 \\
Grammar & Wrong grammar analysis: 41; wrong counting: 5; wrong category: 4 \\
Word formation & Wrong word-formation analysis: 22; wrong counting: 19; wrong category: 9 \\
Spelling & Wrong spelling: 28; wrong counting: 15; wrong category: 15 \\
Others & Wrong language analysis: 26; wrong meaning: 12; wrong counting: 7; wrong factual knowledge: 5 \\
\hline
\end{tabular}
\caption{Manual error analysis of Gemini 2.5 Flash under explained few-shot + justification prompting.}
\label{tab:error_analysis}
\end{table}

\paragraph{Category-level patterns.}
The conceptual categories were dominated by meaning errors. In semantic relatedness and main message, the model often understood the general topic but failed to match the exact intended meaning required by the exam question.

Vocabulary errors were also mostly meaning-related. The model often selected meanings that were semantically plausible but not precise enough for the classical or exam-specific meaning of the word. Some vocabulary errors also involved wrong semantic relations or counting.

In the formal linguistic categories, the dominant problem was incorrect formal analysis. Literary device questions were mainly affected by wrong device identification, while grammar questions were mainly affected by wrong syntactic or grammatical analysis. This suggests that the model lacks precise textbook-style knowledge of Persian literary and grammatical terminology.

The surface-form categories showed a different pattern. In spelling, the model often chose the wrong spelling error or counted the number of errors incorrectly. In word formation, the model often misunderstood how words were built and also made counting mistakes. These categories are difficult because they require very careful attention to letters, word parts, and small details.

\paragraph{Potential solutions.}
The error analysis suggests that different categories require different improvements. For conceptual categories, models may benefit from more Persian literary training data with close distractors and fine-grained explanations. For literary devices and grammar, models need more textbook-style supervision on Persian formal analysis. For spelling and word formation, better prompts alone may not be enough. Future systems could use extra tools that check letters, word parts, and counts more carefully. This could help reduce spelling and counting errors.

\section{Use of Artifacts}
The closed-source models (Gemini 2.5 Flash, GPT-4o, GPT-4.1 Mini, and Grok 4 Fast) were accessed via API under their respective provider terms of service, which permit research use. The open-source models (Llama 3.3 70B Instruct and Qwen2.5 72B Instruct) are released under the Meta Llama 3 Community License and Qwen License, respectively, both of which permit research use. The benchmark dataset is derived from publicly available materials from konkur.in, for which we obtained explicit permission from the website owner for research use and dataset release. Access to PersLitEval will be granted for research purposes only.

\section{AI Usage Statement}
We used AI assistance in several stages of this work. Claude (Anthropic) was used as a writing assistant to help draft, revise, and improve the clarity of the paper text. All AI-generated text was reviewed, edited, and verified by the authors. All experimental results, analyses, and scientific claims are the sole responsibility of the authors.

\end{document}